\pdfoutput=1
\PassOptionsToPackage{table,xcdraw}{xcolor}
\documentclass[11pt]{article}

\usepackage[final]{acl}

\usepackage{times}
\usepackage{latexsym}

\usepackage[T1]{fontenc}

\usepackage[utf8]{inputenc}

\usepackage{float}
\usepackage{amsmath}
\usepackage{amssymb}
\usepackage{booktabs}
\usepackage{subcaption}
\usepackage{diagbox}

\usepackage[most]{tcolorbox} 
\tcbuselibrary{breakable}    

\usepackage{soul}
\sethlcolor{yellow}

\usepackage{xcolor}  

\usepackage{pbox}
\usepackage{makecell}
\usepackage{multirow}
\usepackage{graphicx}

\newcommand*\colourcheck[1]{%
  \expandafter\newcommand\csname #1check\endcsname{\textcolor{#1}{\ding{51}}}%
}
\newcommand*\colouruncheck[1]{%
  \expandafter\newcommand\csname #1uncheck\endcsname{\textcolor{#1}{\ding{53}}}%
}
\colourcheck{green}
\colouruncheck{red}
\colouruncheck{blue}

\newcommand{\update}[1]{\textcolor{black}{#1}}

\usepackage{soul}
\sethlcolor{yellow}

\usepackage{microtype}

\usepackage{inconsolata}

\usepackage{graphicx}

\title{Reassessing Extractive QA Datasets at Scale: LLM-as-a-Judge and In-Depth Analyses
}

\author{
Xanh Ho,$^1$ 
Jiahao Huang,$^2$ 
Florian Boudin,$^{3}$\and
Akiko Aizawa$^{1, 2}$ \\
$^1$National Institute of Informatics, Japan \hspace{1cm}
$^2$The University of Tokyo, Japan \\
$^3$Inria, LS2N, Nantes Université, France \\
{\tt \{xanh, aizawa\}@nii.ac.jp} \hspace{1cm}
{\tt jiahao-huang@g.ecc.u-tokyo.ac.jp} \\
{\tt florian.boudin@univ-nantes.fr} 
}

\begin{document}
\maketitle


\begin{abstract}
Extractive QA tasks are commonly evaluated using Exact Match (EM) and F1-score, but these metrics often fail to reflect true model performance. Recent studies have proposed using large language models (LLMs) as judges (LLM-as-a-judge), yet they often lack comprehensive evaluation across datasets and overlook key factors such as sensitivity to answer types, prompt variations, and self-preference bias.
In this work, we conduct a systematic study of LLM-as-a-judge across four extractive QA datasets and various prompt variations, assessing multiple LLM families in both answering and judging roles. Our results show that LLM-as-a-judge judgments correlate much more strongly with human evaluations than EM (0.22) and F1 (0.40), achieving correlations up to 0.85 with open-source models.
Further analysis reveals that LLM-as-a-judge performs particularly well on number-related answers but faces challenges with more complex types, such as job titles. Contrary to findings in other NLP tasks, we observe no self-preference bias, even when the same model serves as both QA model and judge. Finally, we find that prompt phrasing has minimal impact, and zero-shot, context-free judging often yields the best evaluation performance.\footnote{Our data and code are available at \url{https://github.com/Alab-NII/llm-judge-extract-qa}}
\end{abstract}

\section{Introduction}

Machine reading comprehension (MRC) is a crucial task for evaluating natural language understanding, designed to test a model's reading comprehension by requiring it to answer questions based on a given text~\citep{hirschman-etal-1999-deep}.
Many datasets have been proposed over the past several years, such as SQuAD~\citep{rajpurkar-etal-2016-squad,rajpurkar-etal-2018-know} for simple MRC, QuAC~\citep{choi-etal-2018-quac} and CoQA~\citep{reddy-etal-2019-coqa} for conversational MRC, and QAngaroo~\citep{welbl-etal-2018-constructing},  HotpotQA~\citep{yang-etal-2018-hotpotqa}, and FanOutQA~\citep{zhu-etal-2024-fanoutqa} for multi-hop MRC. 
Based on the answer format, \citet{chen2018neural} classify existing MRC tasks into four types: extraction, multiple-choice (MC), cloze-style, and free-form. 
Some recent datasets also introduce yes/no answers \citep{clark-etal-2019-boolq,geva-etal-2021-aristotle}.

\begin{figure}[t]
\centering
    \includegraphics[width=0.45\textwidth]{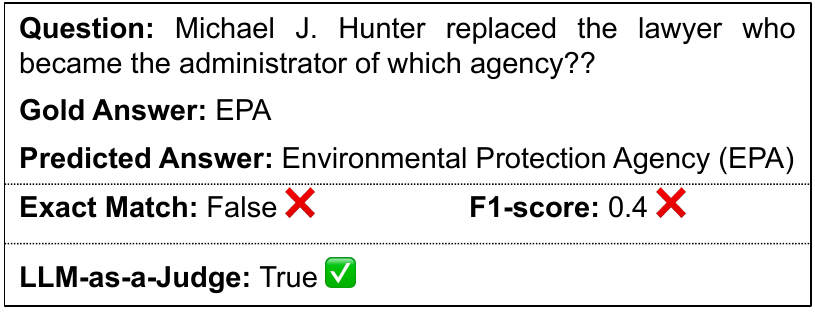}
    \caption{An example that shows EM and F1-score underestimate the performance of models, while LLM-as-a-judge provides a more robust perspective.
    }
    \label{fig_example}
\end{figure}

Depending on the answer type, corresponding evaluation metrics are used. 
While yes/no, multiple-choice, and cloze-style (select-from-options) questions are typically straightforward to evaluate using accuracy as the standard metric, other types of datasets face greater challenges.
For example, extractive QA datasets often rely on Exact Match (EM) and F1-score for evaluation. However, these metrics frequently fail to capture the true performance of models~\cite{risch-etal-2021-semantic,bulian-etal-2022-tomayto}, as they can underestimate correctness when answers are phrased differently but semantically equivalent (as illustrated in Figure~\ref{fig_example}).
This limitation becomes even more pronounced with generative AI models~\cite{kamalloo-etal-2023-evaluating}, which can produce valid answers in a variety of forms. Recognizing this, many recent studies~\cite{kamalloo-etal-2023-evaluating,verga2024replacingjudgesjuriesevaluating,adlakha-etal-2024-evaluating,10.1145/3626772.3657675} have highlighted the inadequacy of traditional metrics and proposed LLM-as-a-judge approaches to provide more reliable and nuanced evaluations of QA systems.

While prior studies have laid the groundwork for using LLMs as judges in QA evaluation, critical aspects of their performance remain unexplored. Specifically, it is unclear how LLM-as-a-judge handles different answer types, whether it exhibits biases, and how robust its judgments are under varying prompting strategies. We argue that systematically analyzing these dimensions is essential for providing more reliable and practical guidance for future research on employing LLM-as-a-judge to evaluate extractive QA answers.
Building on previous research that has explored LLM-as-a-judge for QA evaluation~\citep{kamalloo-etal-2023-evaluating,verga2024replacingjudgesjuriesevaluating,adlakha-etal-2024-evaluating,10.1145/3626772.3657675}, our work extends this line of inquiry through broader experiments and novel analytical perspectives, providing deeper insights into the strengths and limitations of this evaluation paradigm.



Specifically, in this paper, we conduct experiments on four diverse datasets featuring multiple answer types: Quoref~\citep{dasigi-etal-2019-quoref}, DROP~\citep{dua-etal-2019-drop}, HotpotQA~\citep{yang-etal-2018-hotpotqa}, and 2WikiMultiHopQA~\citep{ho-etal-2020-constructing}.
To gain deeper insights into the effectiveness of LLM-as-a-judge for evaluating QA tasks, we analyze the various answer types to identify where LLM-as-a-judge performs well and where it falls short.
Additionally, we employ different families of LLMs both as QA models and as judges, and investigate the presence of self-preference bias~\citep{liu-etal-2024-llms-narcissistic,panickssery2024llm} as well as sibling-preference bias (see Section~\ref{sec_bias_analyze} for details).
%
Finally, to evaluate the robustness and consistency of LLM-as-a-judge, we experiment with various prompt variations, such as changing the number of shots or altering the wording of the prompt.

Our results show that using LLM-as-a-judge correlates highly with human judgments, improving from 0.22 (EM) and 0.40 (F1-score) to 0.85, highlighting its potential to replace these traditional metrics. 
Our analysis demonstrates that LLMs as judges are particularly effective for answer types involving numbers and dates in extractive QA tasks.
Additionally, we observe no evidence of self-preference bias when the same model is used for both QA and judging tasks, nor of sibling-preference bias when a model from the same family serves as the judge.
Finally, changes in both the wording and setup of prompts have little effect on evaluation outcomes, while zero-shot, context-free judging consistently yields the strongest results.

In summary, our work makes four key contributions:
(1) we conduct a comprehensive evaluation of multiple open-weight models across four QA datasets, with code and data released for reproducibility;
(2) we perform the first fine-grained analysis of how judgment quality varies across answer types;
(3)  we investigate the presence of self-preference and sibling-preference biases, finding no evidence of such effects;
(4) we provide a systematic study of prompt robustness, showing that outcomes remain stable across wording and setup variations.
%
Together, our findings offer comprehensive insights that can guide future work on using LLM-as-a-Judge for evaluating extractive or short-form QA datasets.

\section{Related Work}

\paragraph{Extractive QA.}
An extractive QA task requires a model to extract a text span from the provided context to answer a question.
Early datasets for this task include SQuAD~\citep{rajpurkar-etal-2016-squad,rajpurkar-etal-2018-know}, CNN/DailyMail~\citep{NIPS2015_afdec700}, and NewsQA~\citep{trischler-etal-2017-newsqa}. 
Later datasets introduced additional challenges, such as multi-hop reasoning~\citep{welbl-etal-2018-constructing,yang-etal-2018-hotpotqa,trivedi-etal-2022-musique-custom}, coreferential reasoning~\citep{dasigi-etal-2019-quoref}, and numerical reasoning~\citep{dua-etal-2019-drop}.
To the best of our knowledge, previous extractive datasets have primarily relied on automatic evaluation metrics such as EM, F1, or accuracy in their default evaluation protocols.
%

\paragraph{Evaluation Before the LLM-as-a-Judge Era.}
Before the era of LLMs, there were earlier efforts to develop models that assess semantic similarity between candidate answers and gold answers~\cite{chen-etal-2019-evaluating,risch-etal-2021-semantic,bulian-etal-2022-tomayto}.
For example, \citet{bulian-etal-2022-tomayto} introduced BEM (BERT Matching), an automatic metric for evaluating QA performance, and demonstrated that BEM correlates more closely with human judgments than previous metrics such as EM and F1 score.

\paragraph{LLM-as-a-Judge Evaluation.}
Thanks to their success and capabilities, LLMs have been applied to a wide range of tasks, including serving as judges for text generation tasks, such as machine translation~\citep{kocmi-federmann-2023-large}, text summarization~\citep{liu-etal-2023-revisiting,skopek-etal-2023-towards,10.1007/978-3-031-44693-1_54}, story generation~\citep{chiang-lee-2023-large}, and QA~\citep{kamalloo-etal-2023-evaluating,zheng2023judging,verga2024replacingjudgesjuriesevaluating,chen-etal-2024-factors,adlakha-etal-2024-evaluating,10.1145/3626772.3657675}. 

Using LLMs as judges raises several concerns regarding their reliability. 
One common approach to evaluate their performance is by comparing their scores with human judgments to assess correlation~\citep{kamalloo-etal-2023-evaluating,thakur2025judgingjudgesevaluatingalignment}. 
However, prior studies have identified notable biases in LLM-as-a-judge settings, including order bias~\citep{wang-etal-2024-large-language-models-fair} and egocentric bias~\citep{koo-etal-2024-benchmarking}, which can distort evaluation outcomes. 
Additionally, LLMs are vulnerable to adversarial attacks that can manipulate their scoring~\citep{10.1145/3658644.3690291,raina-etal-2024-llm}, further challenging their robustness in judgment tasks.
For a more detailed discussion on the use of LLM-as-a-judge, we refer readers to the comprehensive survey papers by~\citet{li2024llmsasjudgescomprehensivesurveyllmbased} and~\citet{gu2025surveyllmasajudge}.

Unlike previous studies that use LLMs as judges for QA tasks, we conduct a deeper analysis, examining the effects of using LLM-as-a-Judge across different answer types, investigating the presence of self-preference and sibling-preference biases, and assessing the robustness of LLM-as-a-judge under various prompt variations.


\section{Datasets}

We select four reading comprehension QA datasets for our experiments based on the following criteria: (1) they use EM and F1 as evaluation metrics, (2) they have not yet been fully solved, (3) they provide context for the given questions, and (4) they contain diverse types of answers.
These four datasets are: Quoref~\citep{dasigi-etal-2019-quoref}, DROP~\citep{dua-etal-2019-drop}, HotpotQA~\citep{yang-etal-2018-hotpotqa}, and 
2WikiMultiHopQA~\citep[2Wiki;][]{ho-etal-2020-constructing}.
%
To ensure consistency in our comparisons, we only use datasets that employ EM and F1 as evaluation metrics.
To assess whether the current EM and F1 scores may underestimate the true performance of the models, we select datasets where these scores are not excessively high (e.g., below 95\%).
To simplify the judgement process, we choose datasets that provide the corresponding context, allowing us to refer to it when determining whether a predicted answer is acceptable.
Regarding answer types, we select datasets from various tasks, such as numerical reasoning and multi-hop reasoning, to ensure a diverse range of answer types for our analysis.

\paragraph{Quoref.}
Quoref is a reading comprehension dataset designed to assess the coreferential reasoning abilities of models.
In the default dataset setup, a passage can contain multiple QA pairs. 
We treat each pair as an individual sample in our experiments.
The questions are created through crowdsourcing, with a focus on the coreference resolution phenomenon. 
Most of the answer types are in string format, such as person names.

\paragraph{DROP.}
DROP is a reading comprehension dataset designed to evaluate the numerical reasoning abilities of models. 
Similar to Quoref, a single paragraph can contain multiple QA pairs, and we treat each pair as an individual sample.
It is worth noting that, in addition to extractive answer types, DROP also includes number and date answer types. These two types may not always be extractive (i.e., a span of text appearing directly in the context), such as when a question asks about the next day of a specific day.
However, we include them in our analyses, as they do not pose challenges for generative models in predicting the correct answer.

\begin{table*}[t]

  \begin{center}
    \begin{tabular}{l r  r r r r r r | r} 
    \toprule
 

    



   \textbf{Dataset}   & \textbf{String}  & \textbf{Place} & \textbf{Name}  & \textbf{Job}  & \textbf{Date}  & \textbf{Number}  & \textbf{Year}  & \textbf{Total}  \\ \midrule
     
        Quoref & 55 & 74 & 878 & 23 & - & 21 & - & 1,051 \\ 
        DROP & 124 & 24 & 107 & 5 & 18 & 740 & - & 1,018 \\ 
        
        HotpotQA & 146 & 257 & 460 & 92 & 78 & 77 & 14 & 1,124 \\
        
        2Wiki & 296 & 383 & 292 & - & 29 & - & - & 1,000 \\ \midrule
        
        Total & 621 & 738 & 1,737 & 120 & 125 & 838 & 14 & 4,193 \\

    \bottomrule

    \end{tabular}
    \caption{
   The number of samples per answer type and the total number of samples for the four datasets. 
   A dash (`-') indicates that the corresponding answer type is not present in the dataset. 
   In the case of Quoref, we observed that the date and year answer types were frequently mispredicted by rules due to the specific nature of the related questions. 
   Therefore, we chose to exclude these samples from our experiments.
    }
    \label{dataset_info}
  \end{center}
\end{table*}

\paragraph{HotpotQA.}
HotpotQA is a multi-hop QA task that requires multiple reasoning steps to answer each question. 
The dataset comprises two main question types: bridge and comparison. Notably, the presence of comparison questions introduces yes/no answers into the dataset. 
HotpotQA is designed for two tasks: answer prediction and supporting fact prediction. 
In our work, we focus exclusively on the answer prediction task.

\paragraph{2Wiki.}
Similar to HotpotQA, 2Wiki also features two main question types, bridge and comparison. 
In addition, it includes a separate task designed to evaluate the explanatory abilities of models. 
As with HotpotQA, our focus is solely on the answer prediction task.

\paragraph{Obtaining Answer Types.}
From the development set of each dataset, we use heuristic rules to obtain the answer type.
We define 8 answer types as follows: 
\textbf{Place}: questions starting with “where” or asking about locations (city, country, region, etc.); 
\textbf{Name}: questions starting with “who” or “whom” or asking about names, roles, players, actors, etc.; 
\textbf{Job}: questions about occupation, profession, or career; 
\textbf{Date}: questions asking for a specific date; 
\textbf{Number}: questions starting with “how many,” “how much,” or asking about quantities like percentages or populations; 
\textbf{Year}: questions asking for a specific year; 
\textbf{Bool}: yes/no questions;
\textbf{String}: questions that do not fit the other categories.


Our goal is to carefully select a representative subset of approximately 1,000 samples from each dataset to ensure sufficient and meaningful analysis.
It is important to note that our rules may not capture every question from the full development set. 
To ensure diversity in answer types for our experiment, we include all samples from answer types with fewer than 100 instances. 
For answer types exceeding this threshold, we randomly sample in proportion to their representation in the entire dataset.
Since the boolean answer type (yes/no) does not pose significant challenges when evaluated using EM and F1 metrics, we exclude samples with boolean answers from our experiments. 
Table~\ref{dataset_info} presents the number of samples for each answer type, along with the total number of samples for each dataset used in our experiments.

\section{Experimental Settings}

\subsection{Models}

\paragraph{QA Task.}
We use four different model families: Mistral v0.1 (7B and 8x7B)~\citep{jiang2023mistral7b,jiang2024mixtralexperts}, Qwen 2 (7B and 72B)~\citep{yang2024qwen2technicalreport}, Gemma 2 (9B and 27B)~\citep{gemmateam2024gemma2improvingopen},
and Llama 3.1 (8B and 70B)~\citep{grattafiori2024llama3herdmodels}.

\paragraph{LLM-as-a-Judge.}
We use three model families as LLM-as-a-judge systems: Mistral-Instruct-7B-v0.3~\citep{jiang2023mistral7b}, Llama 3.3 70B~\citep{grattafiori2024llama3herdmodels} and Qwen 2.5 72B~\citep{qwen2025qwen25technicalreport}.
It is noted that these model families are similar to those used in our QA task, but we employ an enhanced version. 
To examine the self-preference bias~\citep{liu-etal-2024-llms-narcissistic,panickssery2024llm}, we also run Qwen 2 (7B and 72B) and Llama 3.1 (8B and 70B) as judges in our analyses.
Notably, all models are instruction-tuned.

\begin{table*}[t]
    \centering
    \resizebox{0.7\textwidth}{!}{%
        \begin{tabular}{ l r r r | r r}
    \toprule
   \textbf{QA Model}  &   \textbf{Mistral 7B} & \textbf{Llama 3.3 70B} & \textbf{Qwen 2.5 72B} & \textbf{EM} & \textbf{F1} \\ \midrule

Mistral 7B v0.1 & 0.627 & 0.627 & 0.851 & 0.157 & 0.311 \\
Mixtral 8x7B    & 0.592 & 0.549 & 0.723 & 0.126 & 0.244 \\

Qwen 2 7B       & 0.598 & 0.781 & 0.863 & 0.234 & 0.434  \\
Qwen 2 72B      & 0.653 & 0.680  & 0.793 & 0.157 & 0.274 \\

Gemma 2 9B      & 0.715 & 0.833 & 0.848 & 0.288 & 0.572 \\
Gemma 2 27B     & 0.711 & 0.850  & 0.908 & 0.247  & 0.487 \\

Llama 3.1 8B    & 0.700   & 0.876 & 0.945 & 0.331 & 0.628 \\
Llama 3.1 70B   & 0.626 & 0.803 & 0.848 & \textit{NaN}  & 0.281 \\

\midrule 
Average         & 0.653 & 0.750  & 0.847 & 0.220 & 0.404 \\ 

    \bottomrule
    \end{tabular}
    }
    \caption{
    Pearson correlation coefficients between human judgments and the LLM-as-a-judge evaluations from three models (Mistral 7B v0.3, Llama 3.3 70B, and Qwen 2.5 72B) across eight different QA models are presented. 
    Additionally, we include the correlation scores between human judgments and the EM/F1 scores.
\textit{NaN} values arise when none of the predicted answers by Llama 3.1 70B exactly match the gold answers, resulting in a constant list of 0s, for which correlation is undefined. This often occurs when gold answers are plain numbers (e.g., 93.5), but predictions include extra context like percentages (e.g., 93.5\%) or time spans (e.g., 38 years).
    }
    \label{tab_llm_correlation}
\end{table*}

\subsection{Promptings}
\label{sec:prompt}

\paragraph{QA Task.}
Since most current LLMs are familiar with the QA task and the datasets we use often involve lengthy context passages, coupled with our focus on evaluation, we chose to adopt zero-shot chain-of-thought (CoT) prompting~\cite{NEURIPS2022_8bb0d291} in our experiments.
Our prompt is presented in Appendix~\ref{app_prompt}.

\paragraph{LLM-as-a-Judge.}
To evaluate the predicted answers using LLMs, we use a few-shot prompting approach. 
In this approach, we provide a few demonstration examples to guide the model in labeling the answers. 
Due to length constraints and for the sake of simplicity, these demonstrations exclude the context.
During evaluation, we present the model with a question, a gold answer, a predicted answer, and the corresponding context. 
\update{
%
It should be noted that, while context is essential for QA, it plays a limited role in judging, except in ambiguous cases (e.g., Example \#2 on job type answers in Table~\ref{example_em_f1_llm_judge}).
Nevertheless, we follow the standard setting to report results with context here, while also providing a configuration without context in Section~\ref{sec_prompt_variations}.
}

The model is then instructed to assign one of two labels to the predicted answer: \textbf{CORRECT} (matches the gold answer or a valid alternative) or \textbf{INCORRECT} (does not match the gold answer).
%
We use a 6-shot approach in our judgment prompt. 
We randomly select five samples from each dataset (the subset not used in our evaluation) and run a simple QA model to generate predicted answers.
Subsequently, we manually choose four shots from these samples and reuse two exemplars from the paper by~\citet{verga2024replacingjudgesjuriesevaluating}.
The prompt used in our study is presented in Appendix~\ref{app_prompt}.

\section{Results}
\label{sec:result}

\subsection{Reliability of LLM-as-a-Judge Scores}
\label{human_judge}
To assess the reliability of LLM-as-a-judge scores, we collect human judgments on predicted answers, then calculate the correlation between them and the LLM-as-a-judge scores.

\paragraph{Human Judgements.}

We sampled 200 instances (50 per dataset), maintaining the original distribution of answer types. 
Each sample included all 8 predicted answers. 
Annotators (two authors) were presented with the question, gold answer, predicted answers, and context, and asked to label each predicted answer as either `Correct or `Incorrect'.
Since the task is simple, we decided to assign only one annotator to each sample. However, annotators may discuss ambiguous cases.
Our annotation guideline is presented in Appendix~\ref{app_human_guideline}.
We excluded cases where the gold answer was incorrect, leaving 161 valid samples, each with 8 predicted answers, resulting in 1,288 predicted answers. 
These were used to calculate the correlation between human judgment and LLM performance as a judge.

\begin{table*}[ht!]
  \begin{center}
    \resizebox{0.85\textwidth}{!}{%
    \begin{tabular}{l |  rr >{\columncolor{green!30}} r >{\columncolor{green!30}} r >{\columncolor{cyan!30}}r  rr >{\columncolor{green!30}} r >{\columncolor{green!30}} r >{\columncolor{cyan!30}}r } 
    \toprule
        \multirow{2}{*}{\textbf{Model}} & 
        \multicolumn{5}{c}{\textbf{Quoref}} & \multicolumn{5}{c}{\textbf{DROP}} \\
    \cmidrule(rl){2-6}
    \cmidrule(rl){7-11}

    ~ & EM & F1 &  Mistral &  Llama  &   Qwen  &   EM & F1 & Mistral & Llama &  Qwen   \\ 

\midrule

Mistral 7B v0.1   & 1.2  & 14.9 & 65.6 & 80.5 & 57.9 & 1.6  & 11.3 & 59.8 & 49.4 & 40.9 \\

Mixtral 8x7B & 13.9 & 31.1 & 85.6 & 86.5 & 81.2 & 0.1  & 8.8  & 69.2 & 72.4 & 60.6 \\

Qwen 2 7B         & 33.7 & 46.6 & 76.0   & 64.2 & 61.1 & 18.1 & 27.1 & 65.2 & 54.4 & 48.5 \\

Qwen 2 72B        & 45.5 & 62.3 & 91.8 & 88.8 & 87.1 & 13.9 & 25.6 & 82.1 & 84.6 & 78.3 \\

Gemma 2 9B        & 53.2 & 68.8 & 86.0   & 82.3 & 78.1 & 40.4 & 49.2 & 77.6 & 74.8 & 70.0   \\

Gemma 2 27B       & 58.6 & 72.7 & 88.4 & 82.6 & 80.5 & 45.6 & 56.0   & 80.4 & 80.1 & 75.3 \\

Llama 3.1 8B      & 46.3 & 61.9 & 83.0   & 73.0   & 70.7 & 34.0   & 45.1 & 68.5 & 62.8 & 58.8 \\

Llama 3.1 70B     & 61.0   & 76.9 & 93.8 & 90.6 & 89.7 & 34.2 & 50.8 & 87.3 & 87.4 & 83.3 \\

\midrule
         & 
        \multicolumn{5}{c}{\textbf{HotpotQA}} & \multicolumn{5}{c}{\textbf{2WikiMultihopQA}} \\
    \cmidrule(rl){2-6}
    \cmidrule(rl){7-11}
    
    ~ & EM & F1 & Mistral & Llama  &  Qwen  &   EM & F1 & Mistral & Llama &  Qwen   \\ 

\midrule

Mistral 7B v0.1   & 14.8 & 32.2 & 85.9 & 80.9 & 75.5 & 10.1 & 26.5 & 65.7 & 56.7 & 46.7 \\

Mixtral 8x7B & 7.1  & 26.1 & 85.9 & 89.0   & 84.6 & 6.9  & 25.3 & 73.1 & 73.7 & 66.4 \\

Qwen 2 7B         & 41.2 & 56.9 & 90.8 & 86.3 & 82.4 & 36.4 & 47.9 & 77.3 & 65.9 & 60.4 \\

Qwen 2 72B        & 38.4 & 55.0   & 94.6 & 94.4 & 91.5 & 33.1 & 48.0   & 84.7 & 85.3 & 78.5 \\

Gemma 2 9B        & 56.4 & 72.9 & 94.0   & 91.9 & 88.7 & 52.7 & 61.3 & 77.2 & 77.9 & 68.6 \\

Gemma 2 27B       & 56.6 & 74.3 & 94.7 & 93.1 & 89.9 & 53.1 & 63.8 & 78.9 & 79.5 & 71.8 \\

Llama 3.1 8B      & 51.8 & 68.2 & 91.5 & 89.1 & 85.9 & 45.7 & 55.1 & 80.8 & 71.2 & 63.1 \\

Llama 3.1 70B     & 54.0   & 71.2 & 95.8 & 94.9 & 92.8 & 59.3 & 68.3 & 85.6 & 87.2 & 81.9 \\

    \bottomrule
\end{tabular}

    }
    \caption{
    Automatic evaluation scores (EM and F1) and LLM-as-a-judge scores (highlighted in green and blue) from three models (Mistral 7B v0.3, Llama 3.3 70B, and Qwen 2.5 72B) for the four datasets.
    }
    \label{results_false_samples}
  \end{center}
\end{table*}

\paragraph{Correlation to Human Judgements.}

We calculate the Pearson correlation~\citep{pearson1895regression} between human judgments and the LLM-as-a-judge evaluations from three models: Mistral 7B v0.3, Llama 3.3 70B, and Qwen 2.5 72B. The correlation scores are presented in Table~\ref{tab_llm_correlation}.
As shown in the table, Qwen 2.5 exhibits the highest correlation with human judgments, followed by Llama 3.3, while Mistral v0.3 has the lowest correlation scores. According to standard explanation, a correlation score above 0.80 is generally considered strong.
We focus on using Qwen 2.5 72B as the judge for our subsequent analyses.

Additionally, we report the correlation between human judgments and EM/F1 scores.
\update{
Since F1 is not a binary label, the threshold used for binarization can affect the correlation. We use 0.5 for the main analysis and also tested thresholds of 0.3, 0.4, 0.6, 0.7, and 0.8, yielding average correlations of 0.451, 0.432, 0.379, 0.242, and 0.237, respectively. Even with the most lenient threshold (0.3), the correlation remains only 0.451, lower than all LLM-as-a-judge evaluations (the lowest being 0.653). Full results for all thresholds are provided in Appendix~\ref{app_human_f1}.
}
As shown in Table~\ref{tab_llm_correlation}, EM and F1 scores correlate less with human judgments than any LLM-as-a-judge model. These results suggest that LLMs can serve as reliable judges for evaluating the extractive QA task.

\begin{table}[h]
  \begin{center}
   \resizebox{0.8\columnwidth}{!}{%
                \begin{tabular}{lrr}
        \toprule
        \textbf{Answer Type} & \textbf{\#Samples}& \textbf{Correlation} \\
        \midrule
        name & 464  & 0.862  \\
        number   & 336  & 0.899   \\
        place   & 240   & 0.771  \\
        string   & 160   & 0.862 \\
        job    & 72    & \underline{0.352}     \\
        date     & 16     & \textbf{1.000}  \\                                            \bottomrule
        \end{tabular}
}
    \caption{Correlation scores of Qwen-as-a-judge with human judgment for each answer type.}
     \label{answer_type_corre}
  \end{center}
\end{table}

\subsection{Comparing EM/F1 Scores and LLM-as-a-Judge}
We begin by analyzing the performance gap between EM/F1 scores and LLM-as-a-judge assessments on samples with an EM score of 0.
Next, we compare EM/F1 scores and LLM-as-a-judge across all samples.
Ideally, LLM-as-a-judge should be used only when the EM score is false.

\paragraph{Using LLM-as-a-Judge for False Samples.}
Table~\ref{results_false_samples} presents the EM and F1 scores, along with the LLM-as-a-judge scores (from three models: Mistral 7B v0.3, Llama 3.3 70B, and Qwen 2.5 72B) for the four datasets: Quoref, DROP, HotpotQA, and 2Wiki. 
As shown in the Table, the LLM-as-a-judge scores from the three models are higher than both the EM and F1 scores.
The largest gap is 81.9 between EM and Llama 3.3 70B as a judge on HotpotQA, evaluating Mixtral 8x7B's answers. 
The smallest gap is 15.9 between EM and Qwen 2.5 72B as a judge on 2Wiki, evaluating Gemma 2 9B's answers.
However, this comparison may be biased, as LLM-as-a-judge is only applied to false samples, while samples with a correct EM score are automatically assigned a score of 1.
To offer a broader perspective, we present experimental results for all samples in the following section.

%

\paragraph{Using LLM-as-a-Judge for All Samples.}
Appendix~\ref{app_compare_em_judge} presents LLM-as-a-judge scores for Mistral 7B v0.3, Llama 3.3 70B, and Qwen 2.5 72B across four datasets, evaluated in two cases: false EM samples only and all samples.
As shown in the table, the gaps between these evaluations are minimal, indicating comparable performance.

\paragraph{Summary.}

Table~\ref{results_false_samples} highlights the performance gap between EM/F1 scores and LLM-as-a-judge scores.
Our correlation analysis between human judgments and Qwen-as-a-judge confirms that EM and F1 scores underestimate model performance, not because the LLM-as-a-judge incorrectly labels predictions as correct, but because these metrics fail to capture the true performance.
Notably, using LLM-as-a-judge for false EM samples only or all samples yields similar results.
In practice, we apply LLM-as-a-judge to false EM samples only.

\section{Analyses}

\begin{figure}[t]
    \centering
    \includegraphics[width=0.90\columnwidth]{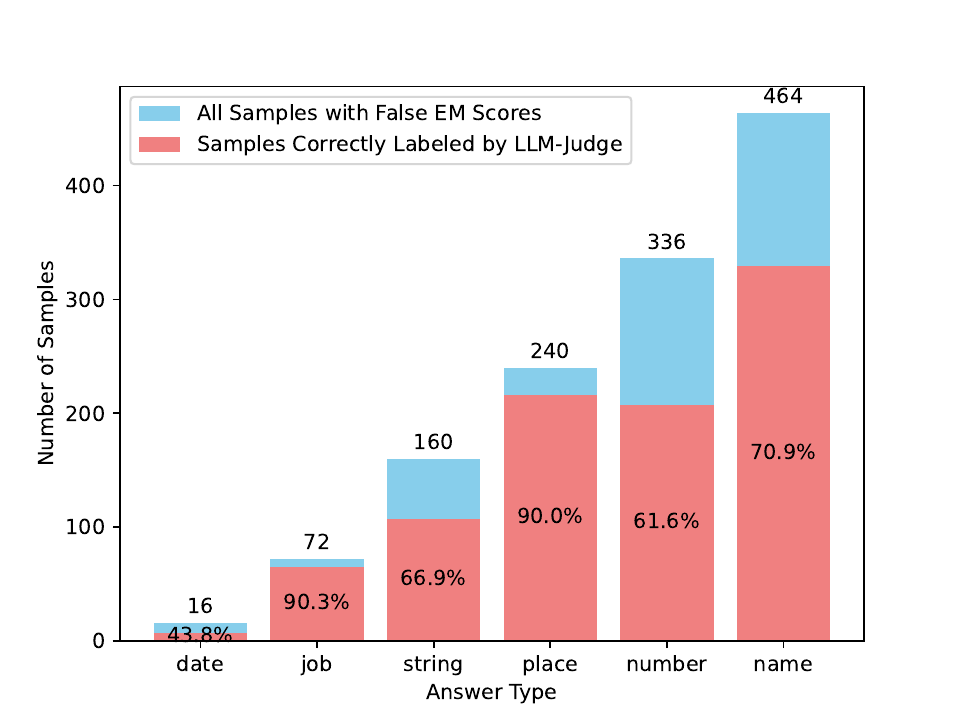}
    \caption{Percentage of each answer type identified as correct by Qwen-as-a-judge.}
    \label{answer_type_percentage}
\end{figure}

\begin{table*}[h]
\centering
 \resizebox{\textwidth}{!}{

\begin{tabular}{l l l l l l p{0.05cm} p{10cm} p{2cm} p{6.7cm} p{1cm} p{2.2cm}}

\toprule

\textbf{\#} & \textbf{Question}  & \textbf{Gold Answer} & \textbf{Predicted Answer} & \textbf{Type} & \textbf{Labels} \\

\midrule
\multirow{3}{0.3cm}{1} 
& \multirow{3}{5.5cm}{What profession did Willi Forst and Elmer Clifton share?} 
& \multirow{3}{2cm}{actor} 
& \multirow{3}{4cm}{Actor, screenwriter, film director, film producer and singer} 
& \multirow{3}{1cm}{Job} 
& \multirow{3}{2.2cm}{\textbf{Human}: F \\ 
\textbf{Model}: T} 
\\ \\ \\

\midrule
\multirow{3}{0.3cm}{2} 
& \multirow{3}{5.5cm}{What is Michael Nakasone job at the college prep school in Honolulu?} 
& \multirow{3}{2cm}{teaches and conducts} 
& \multirow{3}{4cm}{Michael Nakasone is a band director at Punahou School} 
& \multirow{3}{1cm}{Job} 
& \multirow{3}{2.2cm}{\textbf{Human}: F \\ 
\textbf{Model}: T} 
\\ \\ \\

\midrule
\multirow{4}{0.3cm}{3} 
& \multirow{3}{5.5cm}{What was the profession of the one who wrote a song on a 2005 album he collaborated on with Marc Predka?} 
& \multirow{4}{2.5cm}{professional wrestler, actor, and hip hop musician} 
& \multirow{4}{4cm}{hip hop artist} 
& \multirow{4}{1cm}{Job} 
& \multirow{4}{2.2cm}{\textbf{Human}: F \\ 
\textbf{Model}: T} 
\\ \\ \\ \\

\midrule
\multirow{6}{0.3cm}{4} 
& \multirow{6}{5.5cm}{What date did the take over of Enniscorthy end?} 
& \multirow{6}{3cm}{30 April} 
& \multirow{6}{6cm}{The takeover of Enniscorthy ... exact date is not provided in the context. However, since the takeover started on Thursday, 27 April, and lasted until Sunday, we can infer that the takeover ended on 30 April 1916.} 
& \multirow{6}{1cm}{Date} 
& \multirow{6}{2.2cm}{\textbf{Human}: T \\ 
\textbf{Model}: T} 
\\ \\ \\ \\\\ \\

\midrule
\multirow{2}{0.3cm}{5} 
& \multirow{2}{5.5cm}{Why did the director of film The Obsessed Of Catule die?} 
& \multirow{2}{3cm}{stroke} 
& \multirow{2}{6cm}{Anselmo Duarte died due to complications from a stroke} 
& \multirow{2}{1cm}{String} 
& \multirow{2}{2.2cm}{\textbf{Human}: T \\ 
\textbf{Model}: T} 
\\ \\

\midrule
\multirow{2}{0.3cm}{6} 
& \multirow{2}{5.5cm}{How many field goals between 25 and 40 yards were made?} 
& \multirow{2}{3cm}{2} 
& \multirow{2}{5cm}{Two field goals between 25 and 40 yards were made} 
& \multirow{2}{1cm}{Number} 
& \multirow{2}{2.2cm}{\textbf{Human}: T \\ 
\textbf{Model}: T} 
\\ \\

\midrule
\multirow{4}{0.3cm}{7} 
& \multirow{3}{5.5cm}{Who was involved in observing the natural progression of untreated syphilis in rural African-American men in Alabama?} 
& \multirow{4}{2cm}{John Charles Cutler} 
& \multirow{4}{4cm}{The U.S. Public Health Service} 
& \multirow{4}{1cm}{Name} 
& \multirow{4}{2.2cm}{\textbf{Name}: F \\ 
\textbf{Model}: F} 
\\ \\ \\ \\

\midrule
\multirow{3}{0.3cm}{8} 
& \multirow{3}{5.5cm}{Where was the place of burial of Amun-Her-Khepeshef's mother?} 
& \multirow{3}{3cm}{QV66} 
& \multirow{3}{6cm}{The provided text does not mention where Amun-Her-Khepeshef's mother was buried.} 
& \multirow{3}{1cm}{Place} 
& \multirow{3}{2.2cm}{\textbf{Name}: F \\ 
\textbf{Model}: F} 
\\ \\ \\ 





\bottomrule
\end{tabular} 
}
\caption{
Examples of differences and similarities between labels provided by human judgment and LLM-as-a-judge (with labels from Qwen used in this case).
\textit{T} denotes True, while \textit{F} denotes False.
}
\label{example_em_f1_llm_judge}
\end{table*}

\subsection{Answer-Type Effects}

We use 1,288 predicted answers with human judgment in Section~\ref{human_judge} and LLM-as-a-judge scores from Qwen on these samples, and calculate the correlation between the two for each answer type. 
The scores are presented in Table~\ref{answer_type_corre}. 
As expected, the answer types ``date'' and ``number'' show the highest correlation score, while the scores for other types are relatively similar. 
We observe that for the answer type ``job'', the correlation score is quite low, primarily due to the ambiguity of multiple jobs in the gold answer, as the predicted answer can contain more or fewer jobs, making the judgment more difficult. 
Table~\ref{example_em_f1_llm_judge} presents several examples highlighting these differences. 
We also provide additional examples for other answer types.

In Figure~\ref{answer_type_percentage}, we show the percentage of each answer type that LLM-as-a-judge identified as correct when the EM score is false. 
As shown in the figure, the judge is able to correctly predict more than 50\% of each answer type, except for ``date''. 
The highest scores are achieved for the ``job'' and ``place'' answer types. 
However, when considering the correlation score with human judgment for the ``job'' answer type, we observe that LLM-as-a-judge is less strict than humans when judging the correctness of the predicted answer regarding jobs.

\begin{table}[ht]
  \begin{center}
  \resizebox{0.9\columnwidth}{!}{%
\begin{tabular}{l r r r }
\toprule
\textbf{QA Model} & \textbf{Thresh. = 100\%} & \textbf{83\%} & \textbf{67\%} \\
\midrule
Llama 3.1 8B  & 5.77 & 12.04    & 14.77  \\
Llama 3.1 70B & 0.26    & 0.77     & 1.62  \\
Qwen 2 7B     & 0.63    & 2.06    & 4.48  \\
Qwen 2 72B    & 0.14    & 0.34   & 0.85 \\          
\bottomrule
\end{tabular}
}
\caption{Percentage of self-preference bias scores for the four models across three different thresholds.}
\label{self_bias_score}
  \end{center}
\end{table}

\begin{table*}[h]
\centering
\resizebox{\textwidth}{!}{%
\begin{tabular}{ l | c c c c c | c c c}

\textbf{Judge Model} & \textbf{Initial (6-shot)}  & \textbf{Zero-shot} & \textbf{2-shot} & \textbf{No Context} & \textbf{Word Changing} & \textbf{Max} & \textbf{Average} & \textbf{Variance}\\ 
\toprule
Mistral 7B v0.3  &
0.653	& \textbf{0.795}	& 0.667	& 0.762	& 0.702	& 0.795	& 0.716	& 0.0037 \\

Llama 3.3 70B & 

0.750	& 0.819	& 0.742	& \textbf{0.820}	& 0.765& 	0.820 & 	0.779& 	0.0014  \\

Qwen 2.5 72B &  

0.847	 &  0.877	 &  0.850	 &  \textbf{0.909}	 &  0.854	 &  0.909	 &  0.867 &  	0.0007 \\
\bottomrule

\end{tabular}
}
\caption{
Average Pearson correlation coefficients between human judgments and LLM-as-a-judge evaluations across eight different QA models, using the original prompt (as described in Section~\ref{sec:prompt}) and five new prompt variations.
}
\label{tab:qa_prompt_variants}
\end{table*}


\subsection{Self-Preference Bias}
\label{sec_bias_analyze}
We consider four QA models (Llama 3.1 8B, Llama 3.1 70B, Qwen 2 7B, and Qwen 2 72B) and seven LLM-as-a-judge models (Llama 3.1 8B,	Llama 3.1 70B,	Llama 3.3 70B,	Qwen 2 7B,	Qwen 2 72B, Qwen 2.5 72B, and Mistral 7B).
Suppose we have seven judgment models, denoted as $m_1$ to $m_7$ and the QA model is $m_1$.
We define self-preference bias as the scenario in which all remaining models unanimously label the predicted answer as incorrect (threshold = 100\%), while only $m_1$ judges its own answer as correct. 
We also calculate this score for different thresholds, such as 83\%, where one of the six remaining models can judge the answer as correct.
To measure this bias, we calculate the percentage of such cases relative to the total number of cases where the EM score is false.
%

Table~\ref{self_bias_score} presents the percentage of self-preference bias scores for the four models across three different thresholds. 
These scores indicate a relatively small self-preference bias for Llama 3.1 8B when it serves as both a QA model and a judge model. 
However, for the other models, the percentages are much smaller. 
This can be explained by the fact that, in the extractive QA task, where the gold answer is clearly provided, it is more difficult for self-bias to occur compared to other tasks.

We present examples of self-preference bias for different QA models in Appendix~\ref{app_bias}. 
We observe that in these cases, the predicted answer is obviously wrong, and all remaining models label it as incorrect. 
However, only the judge model, which is the same as the QA model, is predicted as correct.

\paragraph{Sibling-Preference Bias.}
\update{
We define sibling-preference bias as a model’s tendency to favor answers generated by another model in the same family. 
Since no self-preference bias was observed, we expect sibling-preference bias to be absent.
%
Using the same four QA models and seven judge models as in the self-preference study, we treat Llama 3.3 70B as a sibling of Llama 3.1 8B/70B and Qwen 2.5 72B as a sibling of Qwen 2 7B/72B.
%
%
Results (Appendix~\ref{app_bias}) confirm that self-preference bias is more pronounced than sibling-preference bias.
}

\subsection{Robustness to Prompt Variations}
\label{sec_prompt_variations}
As discussed in previous works, LLMs may be sensitive to prompt variations~\cite{sclar2024quantifying,voronov-etal-2024-mind,he2024doespromptformattingimpact}. 
To check whether the LLM-as-a-judge is consistent and robust on different types of prompts, as well as to investigate more types of prompts to have a better view for future work, we conduct experiments on other types of promptings, such as changing the configuration to use 2-shot, zero-shot, without context, or changing the words in prompts.
It should be noted that we use greedy decoding for all models.

Table~\ref{tab:qa_prompt_variants} presents the average Pearson correlation coefficients between human judgments and LLM-as-a-judge evaluations across eight different QA models.
The initial (6-shot) prompt is the one described in Section~\ref{sec:prompt}, and its results are reported in Section~\ref{sec:result}.
\textit{Zero-shot} and \textit{2-shot} refer to prompt variations where we change the number of examples (shots) provided.
\textit{No Context} refers to the prompt variation where context paragraphs are excluded for each sample.
\textit{Word Changing} represents the variation where we modify the wording within the prompt.
As shown in the table, zero-shot and no-context prompts often yield the highest scores.
We conjecture that this may be due to the simplicity of the task, which involves evaluating two short answers, making exemplars and additional context less important.
We also observe that the average correlation scores for all three judges are higher than those obtained with the initial prompt.
Moreover, the low variance values across prompt variations (0.0037, 0.0014, and 0.0007) suggest that the LLM-as-a-judge approach is robust and consistently effective, demonstrating strong performance regardless of prompt formulation.

\section{Conclusion}

Our study demonstrates that LLM-as-a-judge offers a more reliable evaluation of extractive QA tasks than traditional metrics such as EM and F1, achieving correlations up to 0.85 with human judgments. The approach performs particularly well on number-related answers but struggles with more complex types, such as job titles. We find no evidence of self-preference bias when the same model is used for both QA and judging, nor sibling-preference bias when the judge belongs to the same model family. Moreover, variations in prompt phrasing or configuration minimally impact results, with zero-shot, context-free judging often yielding the best performance. Overall, our findings highlight that LLMs can serve as reliable, robust evaluators of extractive and short-form QA datasets.

\section*{Limitations}
Our research has three main limitations.
First, we do not conduct experiments using any closed-source LLMs. Since our primary goal is to investigate the effectiveness of using LLM-as-a-judge with open-source models and to support reproducibility for future work, we focus exclusively on open-source LLMs.
Second, we do not evaluate the full versions of the four selected QA datasets. Instead, we conduct experiments on approximately 1,000 samples from each dataset. The exact number of samples used is provided in Table~\ref{dataset_info}.
Third, our human evaluation is limited in scale: only 161 samples were annotated, each with 8 predicted answers from different model outputs, resulting in 1,288 evaluated responses, and each response was assessed by a single annotator.

\section*{Acknowledgments}
This work was supported by JSPS KAKENHI Grant Number 24K03231.

\section*{Licenses}
We use the following datasets for evaluation, in compliance with their respective licenses. HotpotQA (CC BY-SA 4.0) and Quoef (CC BY 4.0) explicitly allow adaptation, redistribution, and modification, while DROP and 2WikiMultihopQA are licensed under the Apache License 2.0, which also permits distribution and modification. We plan to release our dataset under the Apache License 2.0, which allows for redistribution and modification.

\section*{Ethical Statement}

We use publicly available QA datasets in this study. 
To ensure consistency in evaluation, we provide annotators with detailed annotation guidelines and encourage discussion of ambiguous cases during the annotation process. No sensitive or personally identifiable information was collected from annotators at any stage of this research.

\section*{Usage of Large Language Models}
We use large language models (LLMs), such as ChatGPT and GPT-5, for two purposes: (1) improving our writing and grammar, and (2) serving as a QA assistant to identify related works that may be relevant to our research.
It should be noted that all results obtained through these tools are manually verified before being presented in our paper.

\bibliography{anthology,custom}

\clearpage
\newpage 

\appendix

\section{Human Judgements}
\label{app_human_guideline}

Figure~\ref{fig_app_guideline} presents our annotation guideline for evaluating predicted answers against gold answers.

\begin{figure}[ht!]
\centering
\begin{tcolorbox}[colback=cyan!5, colframe=cyan!60, coltitle=black, fonttitle=\bfseries, title=LLM-as-a-Judge - Annotation Guideline, breakable]

\textbf{Input:} \\
- A question \\
- A gold answer \\
- A predicted answer (from an anonymized model) \\
- Context

\vspace{1ex}
\textbf{Output:} \\
- \textbf{Correct (1):} The predicted answer matches the gold answer or is a valid alternative (e.g., different but correct ways of writing a name). \\
- \textbf{Incorrect (0):} The predicted answer is wrong or does not align with the gold answer.

\vspace{1ex}
\textbf{Notice:} In some ambiguous cases, where it is unclear whether the predicted answer is correct or not, please refer to the provided context and use it as the final source for making your judgment.

\end{tcolorbox}
\caption{
Our annotation guideline for evaluating predicted answers against gold answers.
}
\label{fig_app_guideline}
\end{figure}

\section{Experimental Settings}

\begin{table*}[!ht] 
\centering
\resizebox{0.9\textwidth}{!}{%
\begin{tabular}{p{\linewidth}}
\toprule
\textbf{System Prompt}: You are an expert in question answering systems. \\[2ex]

\textbf{User Prompt:} Answer the following question based on the provided context. \\[1ex]

\#\#\# Instructions: \\ 
\textasteriskcentered{} Task: Identify the correct answer from the provided context.  \\
\textasteriskcentered{} Approach: \\ 
- Break down the problem into smaller parts, if necessary. \\ 
- Carefully reason through your answer step-by-step. \\ 
- Ensure that your answer is directly supported by the context. \\[1ex]
\#\#\# Response Format: \\
Please format your answer within brackets as follows: 
\begin{verbatim}
<ans> Your Answer </ans>
\end{verbatim}
\#\#\# \\
\textasteriskcentered{} Question: {\color{blue} \{question\}} \\ 
\textasteriskcentered{} Context: {\color{blue} \{context\}} \\ 

\bottomrule
\end{tabular}
}
\caption{The QA task prompt.}
\label{tab_qa_prompt}
\end{table*}

\subsection{Promptings}
\label{app_prompt}

Table~\ref{tab_qa_prompt} presents the QA task prompt that we used in our experiment.

Table~\ref{tab_llm_judge_prompt} presents the LLM-as-a-judge task prompt used in our experiments.

\begin{table*}[ht]
\centering
\resizebox{\textwidth}{!}{%
\begin{tabular}{p{1\linewidth}}
\toprule
\textbf{System Prompt}: You are an expert in question answering systems. \\[1ex]

\textbf{User Prompt:} Your job is to evaluate a predicted answer by comparing it against the gold answer and the given question. \\ 
You may refer to the provided context if needed. \\[1ex]

\#\# Grading Criteria: \\ 
\textasteriskcentered{} \textbf{CORRECT}: The predicted answer matches the gold answer or is a valid alternative (e.g., different but correct ways of writing a name).  \\
\textasteriskcentered{} \textbf{INCORRECT}: The predicted answer is wrong or does not align with the gold answer. \\ 
\textasteriskcentered{} In some ambiguous cases, where it is unclear whether the predicted answer is correct or not, please refer to the provided context and use it as the final source for making your judgment. \\[1ex]

\#\# Response Format: \\
Please format your answer within brackets as follows: \texttt{<ans> Your Answer </ans>} 

\#\# Here are some examples: \\ [1ex]

\#\#\# Example 1: \\
\textasteriskcentered{} Question: What nationality is the performer of song Daddy, Come Home? \\ 
\textasteriskcentered{} Gold Answer: United States \\ 
\textasteriskcentered{} Predicted Answer: American \\ 
\textasteriskcentered{} Label: \texttt{<ans> CORRECT </ans>} \\ [1ex]

\#\#\# Example 2: \\
\textasteriskcentered{} Question: Who is Bohemond Iv Of Antioch's paternal grandmother? \\ 
\textasteriskcentered{} Gold Answer: Constance of Antioch \\ 
\textasteriskcentered{} Predicted Answer: princess Constance of Antioch \\ 
\textasteriskcentered{} Label: \texttt{<ans> CORRECT </ans>} \\ [1ex]

\#\#\# Example 3: \\
\textasteriskcentered{} Question: Rejuvelac is kind of grain water invented and promoted by a 'holistic health' practitioner born in which year? \\ 
\textasteriskcentered{} Gold Answer: 1909 \\ 
\textasteriskcentered{} Predicted Answer: Rejuvelac is a kind of grain water invented and promoted by Ann Wigmore, who was born in 1909. \\ 
\textasteriskcentered{} Label: \texttt{<ans> CORRECT </ans>} \\ [1ex]

\#\#\# Example 4: \\
\textasteriskcentered{} Question: What is the birthday of the actress who was the Duchess in 'The Revengers Tragedy'? \\ 
\textasteriskcentered{} Gold Answer: 23 November 1946 \\ 
\textasteriskcentered{} Predicted Answer: Diana Quick, who played the Duchess in 'The Revengers Tragedy', was born on 23rd September 1934. \\ 
\textasteriskcentered{} Label: \texttt{<ans> INCORRECT </ans>} \\ [1ex]

(... more examples here ... )



\#\# Here is your task: \\
\textasteriskcentered{} Question: {\color{blue} \{question\}} \\ 
\textasteriskcentered{} Gold Answer: {\color{blue} \{gold\_ans\}} \\ 
\textasteriskcentered{} Predicted Answer: {\color{blue} \{pred\_ans\}} \\ 
\textasteriskcentered{} Context: {\color{blue} \{context\}} \\ 

\bottomrule
\end{tabular}
}
\caption{The LLM-as-a-judge task prompt.}
\label{tab_llm_judge_prompt}
\end{table*}

\section{Results and Analyses}

\subsection{Correlation of Human Judgments and F1}
\label{app_human_f1}

Table~\ref{tab_llm_correlation_f1} shows the Pearson correlation coefficients between human judgments and F1 scores. We use six different thresholds to convert F1 scores into binary labels.

\begin{table*}[t]
    \centering
        \begin{tabular}{ l r r r  r r r }
    \toprule
   \textbf{QA Model}  &   $\mathbf{F_1 \geq 0.3}$ & $\mathbf{F_1 \geq 0.4}$ & $\mathbf{F_1 \geq 0.5}$ & $\mathbf{F_1 \geq 0.6}$ & $\mathbf{F_1 \geq 0.7}$ & $\mathbf{F_1 \geq 0.8}$ \\ \midrule

        Mistral 7B v0.1 & 0.371 & 0.340 & 0.311 & 0.260 & 0.201 & 0.187 \\ 
        
        Mixtral 8x7B & 0.271 & 0.259 & 0.244 & 0.214 & 0.126 & 0.126 \\ 
        
        Qwen 2 7B & 0.487 & 0.459 & 0.434 & 0.404 & 0.224 & 0.224 \\
        
        Qwen 2 72B & 0.339 & 0.309 & 0.274 & 0.254 & 0.148 & 0.148 \\
        
        Gemma 2 9B & 0.595 & 0.597 & 0.572 & 0.563 & 0.375 & 0.369 \\ 
        
        Gemma 2 27B & 0.511 & 0.503 & 0.487 & 0.483 & 0.321 & 0.316 \\ 
        
        Llama 3.1 8B & 0.686 & 0.666 & 0.628 & 0.571 & 0.427 & 0.427 \\
        
        Llama 3.1 70B & 0.351 & 0.320 & 0.281 & 0.285 & 0.110 & 0.102 \\
        
        \midrule 
        Average & 0.451 & 0.432 & 0.404 & 0.379 & 0.242 & 0.237 \\

    \bottomrule
    \end{tabular}
    \caption{
    Pearson correlation coefficients between human judgments and F1 scores. Six different thresholds were used to convert F1 scores into binary labels.
    }
    \label{tab_llm_correlation_f1}
\end{table*}

\subsection{Comparing EM/F1 Scores and LLM-as-a-Judge}
\label{app_compare_em_judge}

Table~\ref{results_all_samples} presents LLM-as-a-judge scores for Mistral 7B v0.3, Llama 3.3 70B, and Qwen 2.5 72B across four datasets.
Columns labeled Mistral, Llama, or Qwen show scores for false EM samples only, while Mistral-A, Llama-A, or Qwen-A show scores for all samples.

\begin{table*}[h]
  \begin{center}
    \resizebox{\textwidth}{!}{%
    \begin{tabular}{l |  rr  rr rr  rr  rr rr } 
    \toprule
        \multirow{2}{*}{\textbf{Model}} & 
        \multicolumn{6}{c}{\textbf{Quoref}} & \multicolumn{6}{c}{\textbf{DROP}} \\
    \cmidrule(rl){2-7}
    \cmidrule(rl){8-13}

    ~ & Mistral & Mistral-A &  Llama &  Llama-A  &   Qwen  &   Qwen-A & Mistral & Mistral-A &  Llama &  Llama-A  &   Qwen  &   Qwen-A   \\ 
\midrule

Mistral 7B v0.1 & 65.6 & 65.7 & 80.5 & 80.6 & 57.9 & 57.6 & 59.8 & 60.2 & 49.4 & 49.8 & 40.9 & 41.0   \\

Mixtral 8x7B    & 85.6 & 86.0   & 86.5 & 87.0   & 81.2 & 82.3 & 69.2 & 72.8 & 72.4 & 75.9 & 60.6 & 62.7 \\

Qwen 2 7B       & 76.0   & 75.9 & 64.2 & 64.2 & 61.1 & 60.8 & 65.2 & 65.2 & 54.4 & 54.4 & 48.5 & 48.0   \\

Qwen 2 72B      & 91.8 & 91.7 & 88.8 & 88.8 & 87.1 & 87.1 & 82.1 & 82.1 & 84.6 & 84.6 & 78.3 & 78.1 \\

Gemma 2 9B      & 86.0   & 85.6 & 82.3 & 82.3 & 78.1 & 77.8 & 77.6 & 77.6 & 74.8 & 74.8 & 70.0   & 70.1 \\
Gemma 2 27B     & 88.4 & 88.0   & 82.6 & 82.6 & 80.5 & 80.4 & 80.4 & 80.4 & 80.1 & 80.1 & 75.3 & 75.6 \\

Llama 3.1 8B    & 83.0   & 82.7 & 73.0   & 72.9 & 70.7 & 70.6 & 68.5 & 68.5 & 62.8 & 62.8 & 58.8 & 58.9 \\

Llama 3.1 70B   & 93.8 & 93.6 & 90.6 & 90.6 & 89.7 & 89.3 & 87.3 & 87.3 & 87.4 & 87.4 & 83.3 & 83.2 \\

\midrule
         & 
        \multicolumn{6}{c}{\textbf{HotpotQA}} & \multicolumn{6}{c}{\textbf{2WikiMultihopQA}} \\
    \cmidrule(rl){2-7}
    \cmidrule(rl){8-13}
    
    ~ & Mistral & Mistral-A &  Llama &  Llama-A  &   Qwen  &   Qwen-A & Mistral & Mistral-A &  Llama &  Llama-A  &   Qwen  &   Qwen-A   \\ 

\midrule

Mistral 7B v0.1 & 85.9 & 85.9 & 80.9 & 80.8 & 75.5 & 75.1 & 65.7 & 65.6 & 56.7 & 56.8 & 46.7 & 46.5 \\

Mixtral 8x7B    & 85.9 & 88.5 & 89.0   & 91.5 & 84.6 & 87.0   & 73.1 & 73.6 & 73.7 & 74.3 & 66.4 & 66.5 \\ 

Qwen 2 7B       & 90.8 & 90.8 & 86.3 & 86.2 & 82.4 & 82.4 & 77.3 & 77.0   & 65.9 & 65.8 & 60.4 & 60.8 \\

Qwen 2 72B      & 94.6 & 94.4 & 94.4 & 94.3 & 91.5 & 92.0   & 84.7 & 84.3 & 85.3 & 85.2 & 78.5 & 78.7 \\

Gemma 2 9B      & 94.0   & 93.8 & 91.9 & 91.8 & 88.7 & 88.3 & 77.2 & 76.6 & 77.9 & 77.8 & 68.6 & 69.0   \\
Gemma 2 27B     & 94.7 & 94.6 & 93.1 & 93.0   & 89.9 & 89.4 & 78.9 & 78.6 & 79.5 & 79.5 & 71.8 & 72.0   \\

Llama 3.1 8B    & 91.5 & 91.4 & 89.1 & 88.7 & 85.9 & 85.8 & 80.8 & 80.5 & 71.2 & 71.2 & 63.1 & 63.2 \\

Llama 3.1 70B   & 95.8 & 95.6 & 94.9 & 94.8 & 92.8 & 92.4 & 85.6 & 85.1 & 87.2 & 87.2 & 81.9 & 81.9 \\
  
    \bottomrule
\end{tabular}

    }
    \caption{
    LLM-as-a-judge scores for Mistral 7B v0.3, Llama 3.3 70B, and Qwen 2.5 72B across four datasets. Columns labeled Mistral, Llama, or Qwen indicate scores on false EM samples only, while Mistral-A, Llama-A, or Qwen-A indicate scores on all samples.
    }
    \label{results_all_samples}
  \end{center}
\end{table*}

\subsection{Self-Preference Bias}
\label{app_bias}

Table~\ref{example_bias} presents examples of self-preference bias for different QA models.

\begin{table*}[h]
\centering
 \resizebox{\textwidth}{!}{
\begin{tabular}{l l l l l l p{0.5cm} p{4cm} p{2cm} p{1cm} p{1cm} p{4cm}}

\toprule

\textbf{Example} & \textbf{Question}  & \textbf{Gold Answer} & \textbf{Predicted Answer} & \textbf{QA Model} & \textbf{Labels} \\

\midrule
\multirow{7}{0.5cm}{1} 
& \multirow{7}{4cm}{What is Abigail's nickname?} 
& \multirow{7}{2cm}{Abby} 
& \multirow{7}{1cm}{Abigail} 
& \multirow{7}{1cm}{Llama 3.1 8B} 
& \multirow{7}{4cm}{judge 1: CORRECT \\ 
      judge 2: INCORRECT \\ 
      judge 3: INCORRECT \\ 
      judge 4: INCORRECT \\
      judge 5: INCORRECT \\
      judge 6: INCORRECT \\
      judge 7: INCORRECT}
\\ \\ \\ \\ \\ \\  \\

\midrule
\multirow{7}{0.5cm}{2} 
& \multirow{7}{4cm}{What was the name of the person that Fauré taught?} 
& \multirow{7}{2cm}{Ravel} 
& \multirow{7}{1cm}{Ralph Vaughan Williams} 
& \multirow{7}{1cm}{Llama 3.1 8B} 
& \multirow{7}{4cm}{judge 1: CORRECT \\ 
      judge 2: INCORRECT \\ 
      judge 3: INCORRECT \\ 
      judge 4: INCORRECT \\
      judge 5: INCORRECT \\
      judge 6: INCORRECT \\
      judge 7: INCORRECT}
\\ \\ \\ \\ \\ \\  \\

\midrule
\multirow{7}{0.5cm}{3} 
& \multirow{7}{4cm}{Where did Mei Shaowu's father die?} 
& \multirow{7}{2cm}{Peking} 
& \multirow{7}{2.7cm}{The context does not provide information about Mei Shaowu's father's death location} 
& \multirow{7}{1cm}{Llama 3.1 70B} 
& \multirow{7}{4cm}{judge 1: INCORRECT \\ 
      judge 2:  CORRECT\\ 
      judge 3: INCORRECT \\ 
      judge 4: INCORRECT \\
      judge 5: INCORRECT \\
      judge 6: INCORRECT \\
      judge 7: INCORRECT}
\\ \\ \\ \\ \\ \\  \\

\midrule
\multirow{7}{0.5cm}{4} 
& \multirow{7}{4cm}{What is the last name of the person who is friends with Cap'n Billau?} 
& \multirow{7}{2cm}{O'Conner} 
& \multirow{7}{2.7cm}{Billau} 
& \multirow{7}{1cm}{Qwen 2 7B} 
& \multirow{7}{4cm}{judge 1: INCORRECT \\ 
      judge 2:  INCORRECT\\ 
      judge 3: INCORRECT \\ 
      judge 4: CORRECT \\
      judge 5: INCORRECT \\
      judge 6: INCORRECT \\
      judge 7: INCORRECT}
\\ \\ \\ \\ \\ \\  \\

\midrule
\multirow{7}{0.5cm}{5} 
& \multirow{7}{4cm}{How many yards longer was the longest field goal compared to the shortest?} 
& \multirow{7}{2cm}{31} 
& \multirow{7}{2.7cm}{18 yards} 
& \multirow{7}{1cm}{Qwen 2 72B} 
& \multirow{7}{4cm}{judge 1: INCORRECT \\ 
      judge 2:  INCORRECT\\ 
      judge 3: INCORRECT \\ 
      judge 4:  INCORRECT\\
      judge 5:  CORRECT\\
      judge 6: INCORRECT \\
      judge 7: INCORRECT}
\\ \\ \\ \\ \\ \\  \\

\bottomrule
\end{tabular} 
}
\caption{
Examples of self-preference bias for different QA models.
}
\label{example_bias}
\end{table*}

Table~\ref{sibling_bias_score} shows the percentage of sibling-preference bias scores for the four models at three different thresholds.

\begin{table}[h]
  \begin{center}
  \resizebox{\columnwidth}{!}{%
\begin{tabular}{l r r r }
\toprule
\textbf{QA Model} & \textbf{Thresh. = 100\%} & \textbf{83\%} & \textbf{67\%} \\
\midrule
Llama 3.1 8B  & 0.60 & 2.47	& 4.39
  \\
Llama 3.1 70B & 0.85 &	2.34 &	3.66  \\
Qwen 2 7B     & 0.03	& 0.30	& 0.83
 \\
Qwen 2 72B    & 0.03 &	0.17 &	0.51
\\          
\bottomrule
\end{tabular}
}
\caption{Percentage of sibling-preference bias scores for the four models across three different thresholds.}
\label{sibling_bias_score}
  \end{center}
\end{table}

\end{document}